%% file: main.tex
\documentclass[10pt, a4paper]{article}
\usepackage{lrec-coling2024} 
\usepackage{natbib}
\usepackage{multibib}
\makeatletter
\def\@mb@citenamelist{cite,citep,citet,citealp,citealt,citepalias,citetalias}
\makeatother
\newcites{languageresource}{~}

\usepackage{graphicx}
\usepackage{tabularx}
\usepackage{soul}
\usepackage{titlesec}
\titleformat{\section}{\normalfont\large\bfseries\center}{\thesection.}{1em}{}
\titleformat{\subsection}{\normalfont\SmallTitleFont\bfseries\raggedright}{\thesubsection.}{1em}{}
\titleformat{\subsubsection}{\normalfont\normalsize\bfseries\raggedright}{\thesubsubsection.}{1em}{}
\renewcommand\thesection{\arabic{section}}
\renewcommand\thesubsection{\thesection.\arabic{subsection}}
\renewcommand\thesubsubsection{\thesubsection.\arabic{subsubsection}}

\usepackage{xcolor}
\usepackage{hyperref}
 \definecolor{darkblue}{rgb}{0, 0, 0.5}
  \hypersetup{colorlinks=true, citecolor=darkblue, linkcolor=darkblue, urlcolor=darkblue}

\usepackage{xstring}

\usepackage{color}

\usepackage[nolist]{acronym}
\input{acronyms}
\usepackage{graphicx}
\usepackage{cleveref}
\usepackage{multirow}
\usepackage{booktabs}
\usepackage{amssymb}
\usepackage{subfigure}
\usepackage[export]{adjustbox}

\title{\vspace*{.5\baselineskip} \textbf{Select and Reorder: A Novel Approach for Neural Sign Language Production}}

\name{Harry Walsh, Ben Saunders, Richard Bowden}

\address{CVSSP, University of Surrey \\
         Guildford, United Kingdom \\
         \{harry.walsh, b.saunders, r.bowden\}@surrey.ac.uk\\}

\abstract{
Sign languages, often categorised as low-resource languages, face significant challenges in achieving accurate translation due to the scarcity of parallel annotated datasets. This paper introduces \acf{snr}, a novel approach that addresses data scarcity by breaking down the translation process into two distinct steps: \acf{gs} and \acf{gr}. Our method leverages large spoken language models and the substantial lexical overlap between source spoken languages and target sign languages to establish an initial alignment. Both steps make use of \ac{nar} decoding for reduced computation and faster inference speeds. Through this disentanglement of tasks, we achieve state-of-the-art BLEU and Rouge scores on the \acf{mdgs} dataset, demonstrating a substantial BLUE-1 improvement of 37.88\% in \acf{ttg} Translation. This innovative approach paves the way for more effective translation models for sign languages, even in resource-constrained settings.
 \\ \newline  }
\begin{document}

\maketitleabstract

\section{Introduction}
\label{sec:intro}
\input{Sections/1_Introduction}

\section{Related Work}
\label{sec:related_word}
\input{Sections/2_RelatedWork}

\section{Methodology}
\label{sec:methodology}
\input{Sections/3_Methology}

\section{Experimental Setup}
\label{sec:setup}
\input{Sections/4_Setup}

\section{Experiments}
\label{sec:experiments}
\input{Sections/5_Experiment}

\section{Conclusion}
\label{sec:conclusion}
\input{Sections/6_Conclusion}

\clearpage

\section{Acknowledgements}
We thank Adam Munder, Mariam Rahmani, and Abolfazl Ravanshad from OmniBridge, an Intel Venture. We also thank Thomas Hanke and the University of Hamburg for the use of the Meine DGS Annotated (mDGS) data.
This work was supported by Intel, the SNSF project ‘SMILE II’ (CRSII5 193686), the European Union’s Horizon2020 programme (‘EASIER’ grant agreement 101016982) and the Innosuisse IICT Flagship (PFFS-21-47). This work reflects only the author's view and the Commission is not responsible for any use that may be made of the information it contains. 

\nocite{*}
\section{Bibliographical References}\label{reference}

\bibliographystyle{lrec-coling2024-natbib}
\bibliography{lrec-coling2024}

\section{Language Resource References}
\label{lr:ref}
\bibliographystylelanguageresource{lrec-coling2024-natbib}
\bibliographylanguageresource{languageresource}

\end{document}

%% file: acronyms.tex
\begin{acronym}[PHOENIX14\textbf{T} ]

\acrodefplural{rnn}[RNNs]{Recurrent Neural Networks}
\acrodefplural{cnn}[CNNs]{Convolutional Neural Networks}
\acrodefplural{hmm}[HMMs]{Hidden Markov Models}
\acrodefplural{gru}[GRUs]{Gated Recurrent Units}
\acrodefplural{crf}[CRFs]{Conditional Random Fields}
\acrodefplural{gan}[GANs]{Generative Adversarial Networks}
\acrodefplural{gpu}[GPUs]{Graphic Processing Units}

\acrodefplural{mdn}[MDNs]{Mixture Density Networks}

\acro{btg}[BTG]{Bracketing Transduction Grammar}
\acro{bpe}[BPE]{Byte Pair Encoding}
\acro{bsl}[BSL]{British Sign Language}
\acro{bleu}[BLEU]{Bilingual Evaluation Understudy}
\acro{blstm}[BLSTM]{Bidirectional Long Short-Term Memory}
\acro{bslcpt}[BSLCP\textbf{T}]{BSL Corpus \textbf{T}}
\acro{cnn}[CNN]{Convolutional Neural Network}
\acro{crf}[CRF]{Conditional Random Field}
\acro{cslr}[CSLR]{Continuous Sign Language Recognition}
\acro{ctc}[CTC]{Connectionist Temporal Classification}
\acro{c4a}[C4A]{Content4All}
\acro{dl}[DL]{Deep Learning}
\acro{dgs}[DGS]{German Sign Language - Deutsche Gebärdensprache}
\acro{dsgs}[DSGS]{Swiss German Sign Language - Deutschschweizer Geb\"ardensprache}
\acro{dtw}[DTW]{Dynamic Time Warping}
\acro{fc}[FC]{Fully Connected}
\acro{ff}[FF]{Feed Forward}
\acro{gan}[GAN]{Generative Adversarial Network}
\acro{gpu}[GPU]{Graphics Processing Unit}
\acro{gru}[GRU]{Gated Recurrent Unit}
\acro{gtpt}[G2PT]{Gloss to Pose Transformer}
\acro{gtp}[G2P]{Gloss to Pose}
\acro{gts}[G2S]{Gloss to Sign}
\acro{gs}[GS]{Gloss Selection}
\acro{gr}[GR]{Gloss Reordering}
\acro{hmm}[HMM]{Hidden Markov Model}
\acro{hpe}[HPE]{Hand Pose Enhancer}
\acro{hoh}[HOH]{Hard of Hearing}
\acro{isl}[ISL]{Irish Sign Language}
\acro{lstm}[LSTM]{Long Short-Term Memory}
\acro{mha}[MHA]{Multi-Headed Attention}
\acro{mtc}[MTC]{Monocular Total Capture}
\acro{mse}[MSE]{Mean Squared Error}
\acro{mdn}[MDN]{Mixture Density Network}
\acro{mdgs}[mDGS]{Meine DGS Annotated}
\acro{mdgst}[mDGS\textbf{T}]{meineDGS\textbf{T}}
\acro{mdgsth}[mDGS\textbf{T}-\textbf{H}]{meineDGS\textbf{T}-\textbf{HARD}}
\acro{mdgste}[mDGS\textbf{T}-\textbf{E}]{meineDGS\textbf{T}-\textbf{EASY}}
\acro{mt}[MT]{Machine Translation}

\acro{nmt}[NMT]{Neural Machine Translation}
\acro{nlp}[NLP]{Natural Language Processing}
\acro{nar}[NAR]{Non-AutoRegressive}
\acro{ph12}[PHOENIX12]{RWTH-PHOENIX-Weather-2012}
\acro{ph14}[PHOENIX14]{RWTH-PHOENIX-Weather-2014}
\acro{ph14t}[PHOENIX14\textbf{T}]{RWTH-PHOENIX-Weather-2014\textbf{T}}
\acro{pof}[POF]{Part Orientation Field}
\acro{pos}[POS]{Part Of Speech}
\acro{paf}[PAF]{Part Affinity Field}
\acro{pttt}[P2TT]{Pose to Text Transformer}
\acro{ptt}[P2T]{Pose to Text}
\acro{ptgtt}[P2G2T]{Pose to Gloss to Text}
\acro{relu}[RELU]{Rectified Linear Units}
\acro{rnn}[RNN]{Recurrent Neural Network}
\acro{rouge}[ROUGE]{Recall-Oriented Understudy for Gisting Evaluation}
\acro{sgd}[SGD]{Stochastic Gradient Descent}
\acro{sla}[SLA]{Sign Language Assessment}
\acro{slr}[SLR]{Sign Language Recognition}
\acro{slt}[SLT]{Sign Language Translation}
\acro{slp}[SLP]{Sign Language Production}
\acro{smt}[SMT]{Statistical Machine Translation}
\acro{slo}[SLO]{Spoken Language Order}
\acro{so}[SO]{Sign Language Order}
\acro{snr}[S\&R]{Select and Reorder}
\acro{sse}[SSE]{Sign Supported English}
\acro{stt}[S2T]{Sign to Text}
\acro{stgtt}[S2G2T]{Sign to Gloss to Text}
\acro{sio}[SIO]{Sign Language Order}
\acro{spo}[SPO]{Spoken Language Order}

\acro{ttgt}[T2GT]{Text to Gloss Transformer}
\acro{ttpt}[T2PT]{Text to Pose Transformer}
\acro{ttp}[T2P]{Text to Pose}
\acro{ttg}[T2G]{Text to Gloss}
\acro{tth}[T2H]{Text to HamNoSys}
\acro{ttgth}[T2G2H]{Text to Gloss to HamNoSys}
\acro{ttgtp}[T2G2P]{Text to Gloss to Pose}
\acro{tts}[T2S]{Text to Speech}
\acro{ttsse}[T2SSE]{Text to Sign Supported English}
\acro{ttspog}[T2SPOG]{Text to Spoken Language Order Gloss}

\acro{wer}[WER]{Word Error Rate}
\acro{wmt14}[WMT2014]{WMT2014 German-English}

\end{acronym}

%% file: Sections/1_Introduction.tex
Sign languages are multi-channel visual languages with complex grammatical rules and structure \citep{stokoe1980sign}. The World Health Organisation estimates that 430 million people worldwide are Deaf or \ac{hoh} \citep{who_2021}, hence the need for accessibility and inclusivity. Sign languages are visual forms of communication, expressed through the manual articulation of gestures and non-manual features. The grammar and lexicon of the world's 300 sign languages are country-dependent and variations can develop from region to region, often sharing a large lexical overlap with each country's respective spoken language \cite{signlanguagenumber}. In the USA, where 90\% of deaf children are born to hearing families  \citep{Schein1974THEDP} sign languages may be acquired at different ages \citep{doi:https://doi.org/10.1002/9780470996522.ch7}, resulting in potential grammar variations \citep{cormier2012first, skotara2012influence}.

\acf{slp} aims to generate sign language sequences from spoken language sentences, it is often decomposed into two concurrent tasks: \acf{ttg}, translating spoken language to gloss sequences, and \ac{gts}, creating sign language videos from gloss intermediaries.  The quality of \ac{slp} videos depends on the initial \ac{ttg} translation. However, current research has predominantly focused on \ac{gts} production \cite{saunders2020adversarial, hwang2021non, huang2021towards, rastgoo2021sign, san2004neural}, leaving a crucial gap in the SLP pipeline. This paper addresses this gap with a novel \acf{snr} approach to \ac{ttg} translation. 
While it is possible to directly synthesise a sign language sequence from a spoken language sentence (\ac{ttp}), a two-step approach has been shown to yield superior translations \citep{saunders2020adversarial}.

\begin{figure*}[htbp!] 
\begin{center}
\includegraphics[width=0.9\textwidth]{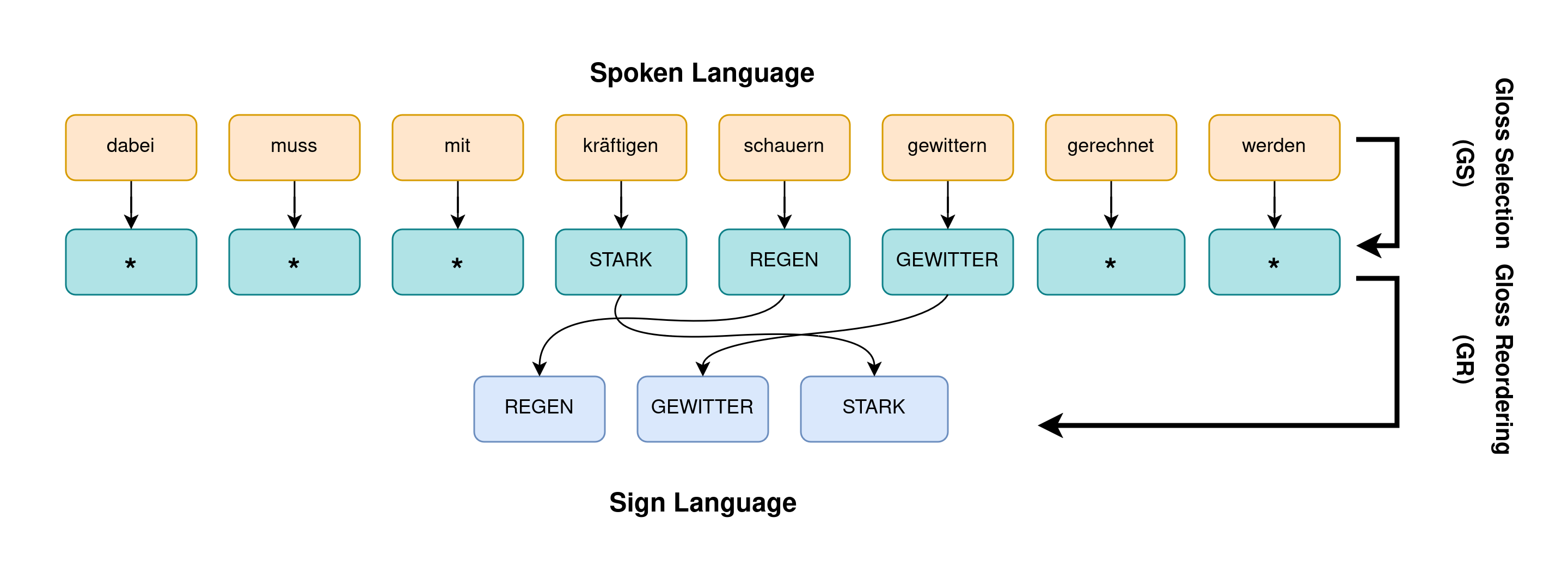}
\caption{An example of \acf{gs} and \acf{gr} being applied to a sentence from the \acf{ph14t} dataset.}
\label{fig:gs_gr_example}
\end{center}
\end{figure*}

To achieve an effective \ac{ttg} translation, it is essential to transform the source spoken language sentence into the target gloss representation while preserving the original meaning. This transformation must include a change in lexicon and in order \citep{el2011towards}, as shown by \Cref{fig:gs_gr_example}. Semantic notations of sign language, such as gloss, share a large proportion of vocabulary with their country of origin. This causes \ac{ttg} translation to have a high lexical overlap between the source and target sequences, a unique property of sign language translation. By first formatting the gloss tokens with lemmatization we find that datasets such as \acf{mdgs} \citetlanguageresource{dgscorpus_3} and \acf{ph14t} \citep{camgoz2018neural} have a lexical overlap of 35\% and 33\%, respectively.

\ac{nmt} typically requires around 15 million sequences of parallel data to outperform statistical approaches \citep{koehn2017six}. By this definition, sign languages can be defined as low-resource languages, with the largest annotated datasets containing only 50k parallel examples \citetlanguageresource{dgscorpus_3}. In an attempt to circumvent this limitation and exploit the lexical overlap, we propose \ac{snr}, an approach that breaks down the translation task into two sub-tasks, \acf{gs} and \acf{gr}. 

As the first step in the \ac{snr} pipeline, \ac{gs} learns to predict the corresponding gloss for each word in the spoken language sentence, thus producing \ac{spo} gloss \citep{marshall2015sign}. To create the ground truth \ac{spo} gloss for training, we obtain a one-to-one alignment between the text and gloss, exploiting the lexical overlap found using large spoken language models such as BERT and Word2Vec.

For the next step, \ac{gr} changes the gloss sequence from \ac{spo} to \ac{sio}. We explore two approaches, a statistical based pre-reordering method \citep{nakagawa-2015-efficient} and a deep learning approach. The statistical approach uses a Top-Down \ac{btg} based pre-ordering model, that learns a number of reordering rules using our alignment, \ac{pos} tags and word classes. The corresponding deep learning approach uses a transformer with a reordering mask at inference time. The mask constrains the model to only predict tokens that are present at the input, meaning the model can only reorder.  

Both \ac{gs} and \ac{gr} are \acf{nar} models, executing decoding in a single pass. This characteristic leads to decreased computational requirements and accelerates the inference process, which is a valuable asset for real-time translation.

The key contributions of this work can be summarized as the following: 
\begin{itemize}
    \itemsep0em 
    \item \ac{snr}, a novel two step approach to \ac{ttg} translation.
    \item An approach to building a pseudo alignment between two paired sequences.
    \item State-of-the-art BLEU and Rouge scores on \ac{mdgs} and \ac{ph14t}.
\end{itemize}

The rest of this paper is organised as follows: In \cref{sec:related_word} we provide an overview of the literature, then in \cref{sec:methodology} we explain our \ac{snr} approach to \ac{ttg} \ac{nmt}. \Cref{sec:setup} explains the setup for the proceeding experiments in \cref{sec:experiments} where we present quantitative and qualitative results. Finally, in \cref{sec:conclusion} we draw conclusions from the experiments and suggest possible future work.

%% file: Sections/2_RelatedWork.tex
\textbf{Sign Language Recognition \& Translation:} For the last 30 years computational sign language Translation has been an active area of research \citep{tamura1988recognition}. Initial neural research focused on isolated \ac{slr}, where \ac{cnn} were used to classify isolated instance of a sign \citep{Lecun_Gradient_based_lrn}. Advancements in the field led to \ac{cslr}, where a video must first be segmented into constituent signs before being classified \citeplanguageresource{KOLLER2015108}. The task of Sign to spoken language translation aims to convert continuous sign language to spoken language text, directly (\ac{stt}) or via gloss (\ac{stgtt}) \citep{camgoz2018neural}.

\textbf{\acf{slp}:} \ac{slp} is the reverse task of \ac{slt},  which aims to produce a continuous sequence of sign language given a spoken language sentence. As above, this can be performed either using gloss as an intermediate representation (\ac{ttgtp}) \citep{stoll2018sign} or directly from the spoken language (\ac{ttp}) \citep{saunders2020adversarial}. State-of-the-art approaches use a transformer with \ac{mha} \cite{saunders2020progressive, stollthere}. The output pose of these systems can be mapped to a photo-realistic signer \citep{saunders2020everybody} or 3D mesh \citep{stollthere}. Older approaches used a parameterized gloss that is converted to a pose and mapped to a graphical avatar \citep{bangham2000virtual, cox2002tessa, zwitserlood2004synthetic, efthimiou2012dicta, van2008virtual}, but this suffers from lack of non-manuals, under-articulation and robotic movement. Recently, alternate representations to gloss have been explored \citep{jiang2022machine, walsh2022changing}, namely SignWriting \citep{kato2008study} and the Hamburg Notation System (HamNoSys) \citep{hanke2004hamnosys}. However, previous work has failed to achieve high \ac{ttg} results, due to the limited dataset size. In this paper, we attempt to overcome the data deficiency by using a \ac{snr} approach.

\textbf{\acf{mt}:} \ac{mt} is an \ac{nlp} task that deals with the automatic translation from a source to a target language. Prior to the introduction of deep learning approaches to the field \citep{singh2017machine}, statistical based methods were state-of-the-art \citep{della1994mathematics, och2002discriminative, koehn2003statistical}. However, these models struggled when the source and target languages had large changes in word order \citep{genzel2010automatically}. To overcome the issues with long-distance word dependencies, pre-reordering was used, where the source language is reordered into the target language order. This was shown to improve the performance of phase based statistical machine translation systems \citep{neubig2012inducing, hitschler2016otedama, nakagawa-2015-efficient}. To train these statistical models an alignment between the source and target words is found \citep{della1994mathematics, vogel1996hmm}. Since then pre-reordering has been applied to \ac{nmt} with limited success \citep{zhao2018exploiting, du2017pre, sabet2020simalign}. Recently word alignment has been used to train multilingual models and has shown good performance when applied to low resource languages \citep{lin2020pre}.

\textbf{Low resource \ac{nmt}:} \ac{nmt} has shown significant performance in large data scenarios but often struggles on low-resource languages \citep{stoll2018sign}. For \ac{slp}, there is a lack of large annotated text to sign corpora. To overcome this, common \ac{nlp} approaches are transfer learning \citep{zoph2016transfer}, use of large language models \citep{zhu2020incorporating} or data augmentation \citep{moryossef2021data}.

%% file: Sections/3_Methology.tex
\acf{ttg} translation aims to learn the mapping from a source spoken language sequence \(X = (x_{1},x_{2},...,x_{W})\) with W words, to a sequence of glosses, \(Y = (y_{1}, y_{2},...,y_{G})\) with G glosses. Therefore, a \ac{ttg} model learns the conditional probabilities \(p(Y|X)\). 

\begin{figure}[!ht]
    \centering
    \includegraphics[width=0.4\textwidth]{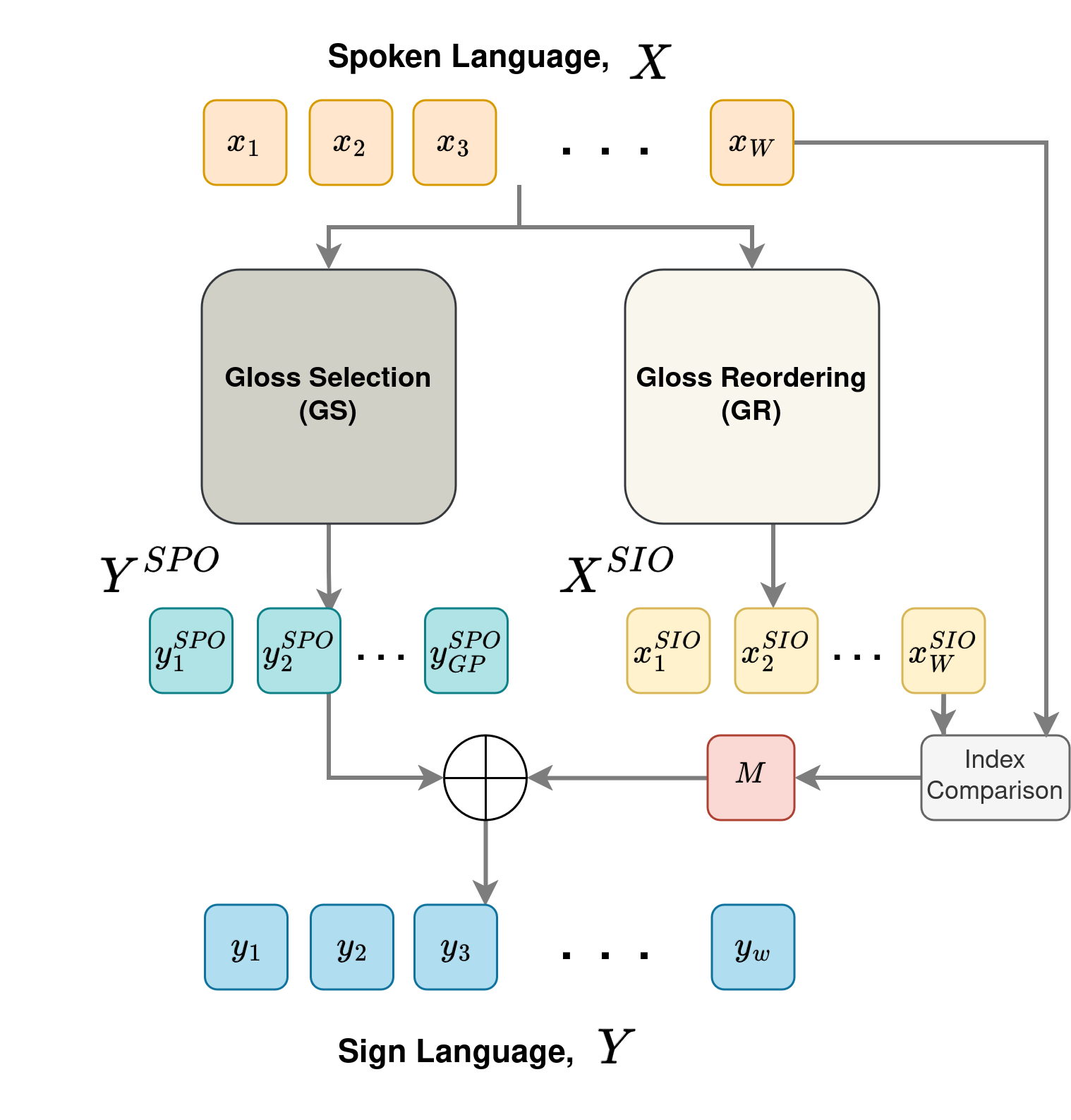}
    \caption{A overview of the \acf{snr} approach}
    \label{fig:architecture}
\end{figure}

A model that learns \(p(Y|X)\) jointly learns a change in lexicon and order, a challenging task. In this paper, we disentangle the two tasks into \ac{gs} and \ac{gr}, as shown in \Cref{fig:architecture}, and define a new task of \ac{ttspog}.

The \ac{gs} model learns the mapping of a sequence of words \(X = (x_{1},x_{2},...,x_{W})\), to a sequence of \ac{spo} glosses and pad tokens, \(Y^{SPO} = (y^{SPO}_{1}, y^{SPO}_{2},...,y^{SPO}_{W})\). Our approach relies on creating a one-to-one alignment, \(A\), of words to glosses, which limits us to sequences where (\(W \ge G\)), hence \(X\) and \(Y^{SPO}\) share the same sequence length, W. To create the gloss in \ac{spo} for the \ac{gs} model the alignment, \(A()\), is applied to the gloss;
\begin{equation} Y^{SPO} = A(Y) \end{equation}

We define \ac{gr} as a permutation task, where the model learns to reorder words in spoken language order, \(X\), to words in sign order, \(X^{SIO} = x^{SIO}_{1}, x^{SIO}_{2},...,x^{SIO}_{W})\). The source and target sequence share the same vocabulary and sequence length, W. Thus, \ac{gr} learns \(p(X^{SIO}|X)\). To create the text in sign order for the \ac{gr} model the alignment, \(A()\), is applied to the text;
\begin{equation} X^{SIO} = A(X) \end{equation}

As shown by \Cref{fig:architecture}, to obtain a full translation the outputs of the \ac{gs} and the \ac{gr} networks must be joined. We call this full method \acf{snr}. To correctly join the outputs the \ac{gr} subtask creates a mapping \(M()\). Applying the mapping to the \ac{spo} gloss gives a full translation (gloss in sign language order);
\begin{equation} p(Y|X) = M(p(Y^{SPO}|X))  \end{equation}

Both input and target sequences are tokenized at the word level. The \ac{gs} and \ac{gr} networks are trained using cross-entropy loss, \(L_{cross}\), calculated using the predicted target sequence, \(\hat{x}\) and the ground truth sequence, \({x}^{*}\).  

In the following sub-sections, we provide an overview of \ac{gs} followed by \ac{gr}. Firstly, we show how we create an alignment between the source and target languages, using two different word embeddings. Subsequently, we explain how this alignment is used in conjunction with the \ac{gs} model to predict the intermediary \ac{spo} glosses. Finally, we outline two methods for \ac{gr}, followed by an explanation of how the two sub-tasks are joined to obtain a full translation.

\subsection{Select}
\label{sec:select}

As shown by \Cref{fig:gs_gr_example} (top to middle row), \ac{gs} can be defined as the task of choosing the corresponding glosses for each word of a given spoken language sentence. To achieve this, an alignment must first be found to create a pseudo gloss sequence in \ac{spo}.

\subsubsection{Alignment}
\label{sec:alignment}
Using the lexical overlap between the source and target language, a pseudo alignment can be found. For example given the sentence "\textit{what is your name?}" it is clear to see which words correspond to which glosses in the translation \textit{"YOU NAME WHAT?"}. Using two different word embedding techniques, Word2Vec \citep{mikolov2013distributed} and BERT \citep{chan-etal-2020-germans}, we create a mapping between our spoken language words, X, and our glosses, Y. We can define a word gloss pair as a strong alignment if they share the same meaning. A strong connection can be established if the pair share a similar lexical form (e.g. word = run, gloss = RUN), for which we use Word2Vec. Where an accurate lexical mapping cannot be found, we use BERT to find connections based on meaning (e.g. word = weather, gloss = WEATHERFORECAST). When using \ac{dgs} we first apply a compound word splitting algorithm \citep{tuggener2016incremental} before creating the alignment.

For a sequence of words, \(X\), and a sequence of gloss, \(Y\) we apply Word2Vec as: \begin{equation} X_{Vec} = Word2Vec(X) \end{equation}
\begin{equation} Y_{Vec} = Word2Vec(Y) \end{equation}
where $ X_{Vec} \in \mathbb{R}^{W \times E}$ and 
$ Y_{Vec} \in \mathbb{R}^{G \times E}$. 
We take the outer product between the resultant two embeddings to give us the Word2Vec alignment: 
\begin{equation} A_{Vec} = Y_{Vec} \otimes X_{Vec} \end{equation}
where $ A_{Vec} \in \mathbb{R}^{G \times W}$. 
We filter the strongest connections, only keeping those that are above a constant, $\alpha$. Then we repeat the process this time using BERT:
\begin{equation} X_{BERT} = BERT(x) \end{equation}
\begin{equation} Y_{BERT} = BERT(y) \end{equation}
\begin{equation} A_{BERT} = Y_{BERT} \otimes X_{BERT} \end{equation}
Where $ X_{BERT} \in \mathbb{R}^{W \times E}$, $ Y_{BERT} \in \mathbb{R}^{G \times E}$ and $ A_{BERT} \in \mathbb{R}^{G \times W}$. When embedding with BERT, a wordpiece tokenizer is applied to the text. We average the sub-unit alignment in order to create an alignment at the word level. We find BERT embeddings capture the meaning of tokens, making this approach better for finding alignment between words and glosses that have different lexical forms. The BERT alignment is used to find any remaining connections not found by the Word2Vec alignment, where our final alignment is defined as;
\begin{equation} A = A_{BERT} + (\alpha * A_{Vec}) \end{equation}

Note, as the alignment creates a one-to-one mapping, the proposed approach is limited to many-to-one sequences e.g. where the source sequence is longer than the target, (\(W \ge G\)). Furthermore, as the \ac{ttspog} task is a many-to-one task, any words that are not aligned are mapped to a pad token, '*'. This ensures the sequence lengths of \(Y^{SPO}\) and \(X\) are the same.

\begin{figure}[!ht]
    \centering
    \includegraphics[width=0.46\textwidth]{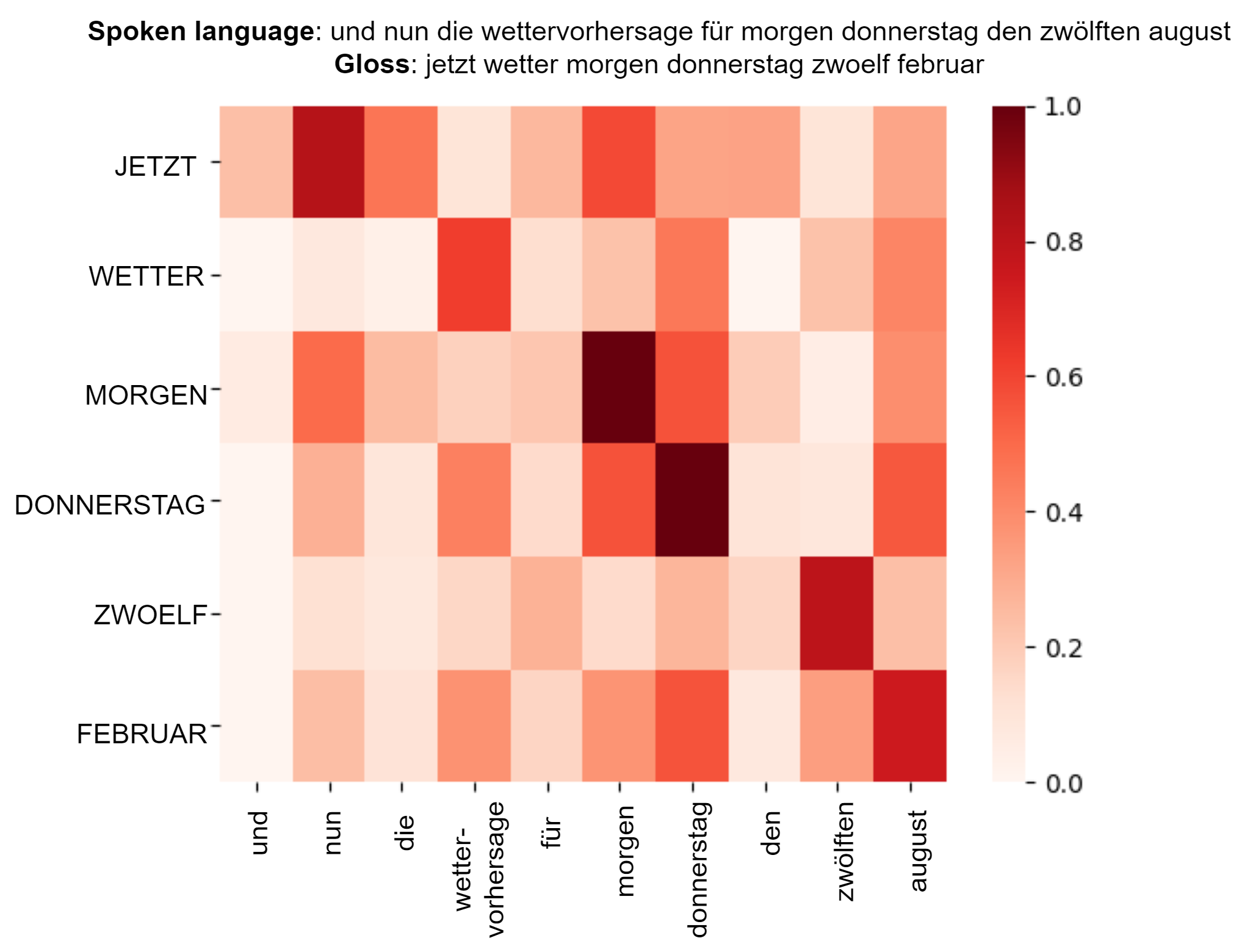}
    \caption{An example of the alignment found using BERT embeddings to connect the spoken language to the glosses on the \ac{ph14t} dataset. (SRC: "and now the weather forecast for tomorrow thursday the twelfth of august", TRG: "now weather tomorrow thursday twelve february")}
    \label{fig:phix_alignement_heatmap}
\end{figure}

\begin{figure}[!ht]
    \centering
    \includegraphics[width=0.46\textwidth]{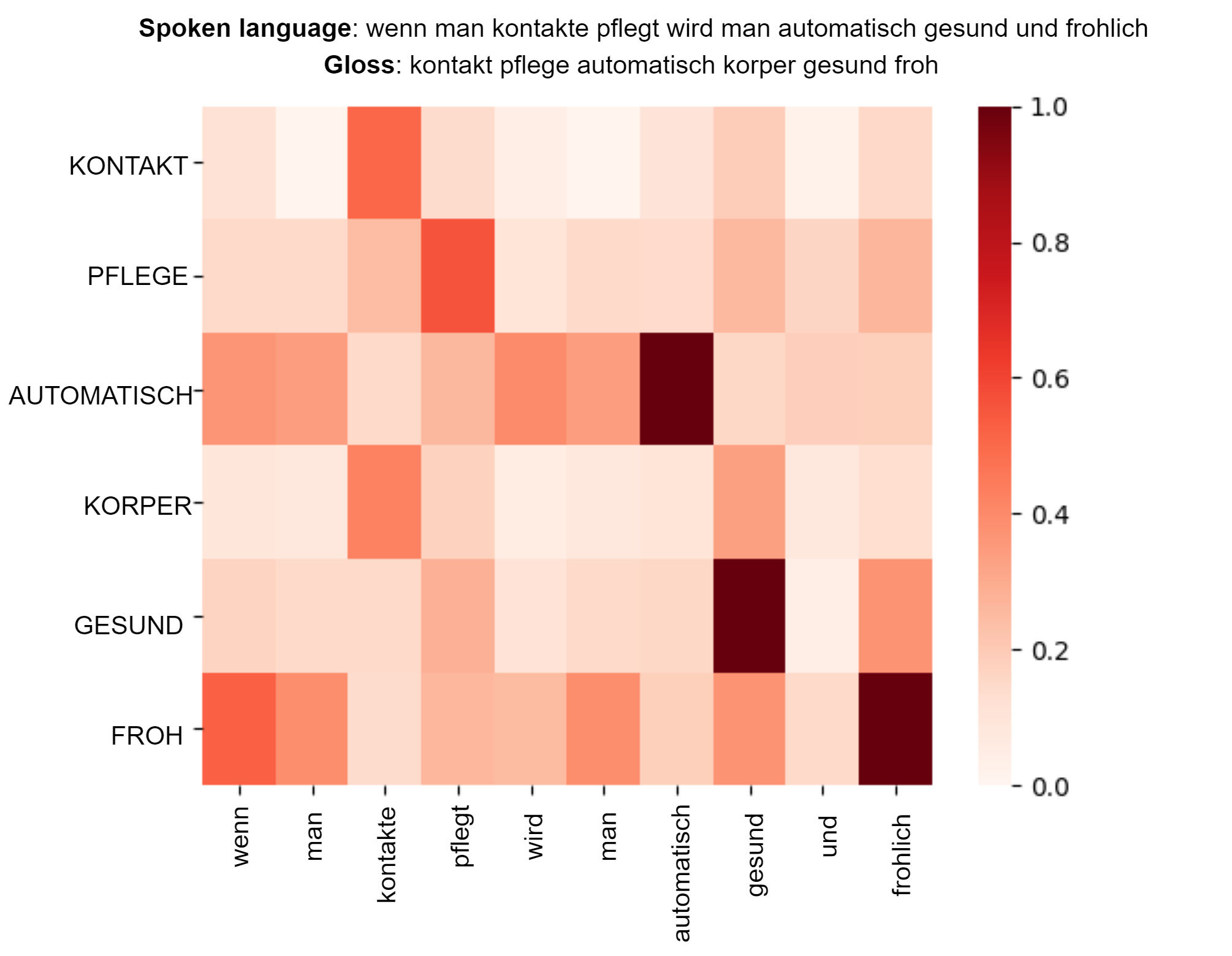}
    \caption{An example of the alignment found using BERT embeddings to connect the spoken language to the glosses on the \ac{mdgs} dataset. (SCR: "when you keep in touch you automatically become healthy and happy", TRG: "contact care automatic body healthy glad")}
    \label{fig:mdgs_alignement_heatmap}
\end{figure}

\Cref{fig:phix_alignement_heatmap} and \ref{fig:mdgs_alignement_heatmap} shows a heat map of the alignment found between a German spoken language sentence and the corresponding gloss sequence from the \ac{ph14t} and \ac{mdgs} dataset, respectively. \Cref{fig:phix_alignement_heatmap} shows a clear alignment is found between the word and gloss "MORGEN", as they share the same lexical form. Additionally, an alignment is found between words with the same meaning e.g. "JETZT" and "nun". 

\subsubsection{Architecture}

We build our \ac{gs} model as an encoder-decoder transformer \citep{vaswani2017attention}. We pass the encoder and decoder the same spoken language sentence, whilst removing the auto-regressive feature of the decoder for reduced computational cost. This also makes each prediction independent of the previous, removing any possible negative feedback from incorrect guesses at inference time. Additionally, we alter the decoder's forward masking to allow the model to see all tokens in the sequence.

\subsection{Reorder}

The goal of the second sub-task, \ac{gr}, is to create a mapping, \(M()\), that reorders a sequence from \ac{spo} to \ac{sio}. By comparing the index movements of words between the input and output of the models we create a mapping. A visualization of applying \(M()\) can be seen in \Cref{fig:gs_gr_example} (middle to bottom row).

To facilitate the creation of ground-truth data for this task, we can leverage the alignment discussed in \cref{sec:alignment}, referred to as  \(A\). This alignment enables us to generate \ac{spo} gloss and text in \ac{sio}. Consequently, we have the option to train our reordering model on either the gloss or the text. We opt to train on the text for two reasons. Firstly, gloss does not offer a perfect representation of sign language due to its inherent limitations. Secondly, we hypothesise that training on the higher-resourced language (text) will yield superior performance, as it contains richer structural information about the language.

In this section, we propose two approaches to tackle \ac{gr}. We start by explaining the statistical approach from \citet{nakagawa-2015-efficient}, followed by our deep learning method.

\subsubsection{Statistical Approach}

Our first approach uses the \ac{btg} method to learn a mapping from spoken to sign order \citep{nakagawa-2015-efficient}. The approach represents a source sentence as a binary tree, where each non terminal node can be one of three types: straight, inverted or terminal. The structure of the tree is dependent on the \ac{pos} tags and word classes of the sentence. 
To create our word classes we use the Brown clustering method \cite{brown1992class}. Words are grouped into a single cluster if they are semantically related. Words are assumed to be semantically related if the distribution of surrounding words are similar. We use a pre-trained language model to tag the spoken language words with their \ac{pos}.

The model acquires a set of rules designed to restructure the tree in a manner that maximises reordering accuracy. To assess this accuracy, we rely on the alignment provided in \cref{sec:alignment}.

\input{Tables/Gloss_Selection_results}

\subsubsection{Learned Approach}

Our second approach uses deep learning to learn \(p(X^{SIO}|X)\), using an encoder-decoder transformer. Once again we remove the auto-regressive feature from the decoder and change the mask to allow the model to see all tokens in the input sequence. At inference time we apply a mask to the output of the model, which ensures the model predicts all tokens that are present on the input, hence the model is limited to reordering. The mask is a binary vector with entries only in the index's of the tokens present in the input. At each decoding step, the predicted token is removed from the mask. If duplicates of the gloss are present then it is only removed once all copies have been predicted. 

\subsection{Select and Reorder}

The \ac{gs} model learns \(p(Y^{SPO}|X)\) and the \ac{gr} model learns \(p(X^{SIO}|X)\). To obtain a full translation the output of the two models must be joined, as shown by \Cref{fig:architecture}. As depicted, the predictions of the \ac{gr} cannot be directly applied to the gloss in \ac{spo}. By analysing the index movement of words between the input and output a mapping, \(M\), can be created that changes the order from spoken to sign. We train each task independently and join the outputs by applying the mapping, \(M()\) from \ac{gr} to the output of the \ac{gs} model;
\begin{equation} Y = M(Y_{SPO}) \end{equation} 
This provides a full translation from a spoken language input sequence to a target gloss sequence.

%% file: Tables/Gloss_Selection_results.tex
\begin{table*}[ht]
\centering
\resizebox{0.9\linewidth}{!}{%
\begin{tabular}{r|ccccc|ccccc} 
\toprule
\multicolumn{1}{l}{\ac{ph14t}}     & \multicolumn{5}{c}{DEV SET}               & \multicolumn{5}{c}{TEST SET}               \\
\multicolumn{1}{c|}{} & BLEU-1 & BLEU-2 & BLEU-3 & BLEU-4 & ROUGE & BLEU-1 & BLEU-2 & BLEU-3 & BLEU-4 & ROUGE  \\ 
\midrule
GS (Spoken order)                  & 62.69  & 41.22  & 29.04  & 21.31  & 58.32 & 60.12  & 39.22  & 27.40  & 20.19  & 57.10  \\ 
GS (Sign order)                           & 62.69  & 38.86  & 25.67  & 17.84  & 56.37 & 60.13  & 35.15  & 21.84  & 14.49  & 54.60  \\
\bottomrule
\end{tabular}
}
\caption{\label{tab:PHIX_gloss_selection}A table showing the result of performing \acf{gs} on the \acf{ph14t} dataset.}
\end{table*}

\begin{table*}[ht]
\centering
\resizebox{0.9\linewidth}{!}{%

\begin{tabular}{r|ccccc|ccccc} 
\toprule
\multicolumn{1}{l}{\ac{mdgs}}      & \multicolumn{5}{c}{DEV SET}               & \multicolumn{5}{c}{TEST SET}               \\
\multicolumn{1}{c|}{} & BLEU-1 & BLEU-2 & BLEU-3 & BLEU-4 & ROUGE & BLEU-1 & BLEU-2 & BLEU-3 & BLEU-4 & ROUGE  \\ 
\midrule
GS (Spoken order)                  & 42.91  & 23.21  & 12.47  & 6.89   & 42.63 & 43.06  & 23.23  & 12.71  & 7.02   & 42.65  \\ 
GS (Sign order)                           & 42.91  & 20.51  & 9.86   & 4.95   & 40.63 & 43.06  & 20.97  & 10.48  & 5.39   & 40.60  \\
\bottomrule
\end{tabular}
}
\caption{\label{tab:MDGS_gloss_selection}A table showing the result of performing \acf{gs} on the \acf{mdgs} dataset.}
\end{table*}

%% file: Sections/4_Setup.tex
In this section, we explain the experimental setup for the proceeding experiments. 

To initialize the encoder and decoder of the transformer we use xavier initializer \citep{glorot2010understanding} with zero bias and Adam optimization \citep{kingma2014adam}. The initial learning rate is set to \(10^{-4}\) with a decrease factor of 0.7 and patience of 5. During training we employ dropout connections, therefore we apply a dropout probability of 0.35 and 0.2 for the \ac{gs} and \ac{gr} models respectively \citep{srivastava2014dropout}. When decoding we apply a greedy algorithm on both models. We filter the confidence of the word2vec alignment, \(A_{Vec}\), with a factor of 0.9. We train the \ac{btg} prereorder model for 30 iterations with a beam size of 20 on the training set only. We set the number of word classes to 50 when clustering with \citet{brown1992class} and we tag the spoken language with \ac{pos} using the Spacy python implementation for German. 

Our code base comes from the \citet{kreutzer2019joey} NMT toolkit, JoeyNMT \citep{kreutzer2019joey} and is implemented using Pytorch \citep{NEURIPS2019_9015}. When embedding with BERT, we use an open source pre-trained model from Deepset \citep{chan-etal-2020-germans}. Finally, we used fasttext's implementation of Word2Vec for word level embedding \citep{mikolov2013distributed}. 

To evaluate our models, we use the Public Corpus of German Sign Language, 3rd release, the \ac{mdgs} dataset \citeplanguageresource{dgscorpus_3} and the \ac{ph14t} dataset \citep{camgoz2018neural}. \ac{mdgs} contains aligned spoken German sentences and gloss sequences, from unconstrained dialogue between two native deaf signers \citeplanguageresource{dgscorpus_3} and we use the translation protocol set in \citet{saunders2021signing}.

\ac{mdgs} is 7.5 times larger compared to \ac{ph14t} with 330 deaf participants performing free-form signing and a source vocabulary of 18,457. Note we remove the gloss variant numbers to reduce singletons. We use BLEU scores (BLEU-1,2,3 and 4) and Rouge score to evaluate all methods.

\input{Tables/Gloss_Reordering_results}

%% file: Tables/Gloss_Reordering_results.tex
\begin{table*}
\centering
\resizebox{0.9\linewidth}{!}{%
\begin{tabular}{r|ccccc|ccccc} 
\toprule
\multicolumn{1}{l}{\ac{ph14t}}     & \multicolumn{5}{c}{DEV SET}               & \multicolumn{5}{c}{TEST SET}          \\
\multicolumn{1}{c|}{Mapping:} & BLEU-1 & BLEU-2 & BLEU-3 & BLEU-4 & ROUGE & BLEU-1 & BLEU-2 & BLEU-3 & BLEU-4 & ROUGE  \\ 
\midrule
GT (Aligned Gloss)            & 100.00 & 66.36  & 48.72  & 37.43  & 76.21 & 100.00 & 64.00  & 45.60  & 34.07  & 76.17  \\ 
Learnt                        & 99.14  & 60.14  & 39.50  & 26.93  & 59.17 & 99.20  & 59.43  & 38.58  & 25.74  & 58.28  \\
Statistical                   & 100.00 & 76.52  & 62.83  & 53.44  & 84.61 & 100.00 & 61.87  & 42.64  & 31.05  & 74.66  \\

\bottomrule
\end{tabular}
}
\caption{\label{tab:PHIX_reordering}A table showing the results of performing \acf{gr} from \acf{spo} to \acf{sio} on the \acf{ph14t} dataset.}
\end{table*}

\begin{table*}
\centering
\resizebox{0.9\linewidth}{!}{%
\begin{tabular}{r|ccccc|ccccc} 
\toprule
\multicolumn{1}{l}{\ac{mdgs}}      & \multicolumn{5}{c}{DEV SET}               & \multicolumn{5}{c}{TEST SET}          \\
\multicolumn{1}{c|}{Mapping:} & BLEU-1 & BLEU-2 & BLEU-3 & BLEU-4 & ROUGE & BLEU-1 & BLEU-2 & BLEU-3 & BLEU-4 & ROUGE  \\ 
\midrule
GT (Aligned Gloss)            & 100.00 & 65.20  & 43.72  & 29.89  & 80.20 & 100.00 & 64.69  & 42.98  & 29.35  & 80.37  \\ 
Learnt                        & 97.62  & 59.45  & 40.40  & 29.29  & 60.75 & 97.60  & 59.36  & 40.36  & 29.33  & 60.32  \\
Statistical                   & 100.00 & 82.12  & 68.67  & 57.93  & 91.24 & 100.00 & 60.06  & 36.87  & 22.91  & 77.47  \\ 

\bottomrule
\end{tabular}
}
\caption{\label{tab:MDGS_reordering}A table showing the results of performing \acf{gr} from \acf{spo} to \acf{sio} on the \acf{mdgs} dataset.}
\end{table*}

%% file: Sections/5_Experiment.tex
\subsection{Quantitative Experiments}

In this section, we evaluate our proposed approaches on the \ac{mdgs} and the \ac{ph14t} dataset. We group our experiments in four sections:
\begin{enumerate}
    \itemsep0em 
    \item \acf{gs}.
    \item \acf{gr}.
    \item \ac{snr} (\ac{gs} \(+\) \ac{gr}) and State-of-the-art Comparison. 
    \item Inference speed tests. 
 \end{enumerate}

\subsubsection{Gloss Selection}

Firstly, we evaluate our \ac{gs} approach. As discussed in \cref{sec:select} we create an alignment for both datasets in order to perform \ac{gs}. \Cref{tab:PHIX_gloss_selection} and \ref{tab:MDGS_gloss_selection} show the results. In both cases, the GS output is the same but compared against the \ac{spo} (row 1) and the ground truth gloss (\ac{sio}) (row 2), hence the BLEU-1 score is the same. As the model was trained to predict \ac{spo} order, it is not surprising that the BLEU-4 score is higher. However, the performance drop when evaluated against sign order is small. The high BLUE-1 scores demonstrate the effectiveness of this method, achieving 42.91 on the challenging \ac{mdgs} dataset.

\subsubsection{Gloss Reordering}
\label{sec:reodering}
Next, we compare our different reordering approaches. When evaluating the \ac{gr} model any words not present in the training set are replaced with unknown tokens. Thus, the BLEU score for the learnt method is not 100, even though the model has to predict all words that are present at the input.  As can seen from \cref{tab:PHIX_reordering} and \ref{tab:MDGS_reordering}, the statistical method outperforms the learnt approach, with the statistical method achieving 26.51 and 28.64 BLEU-4 higher on the \ac{ph14t} and \ac{mdgs} dev sets respectively. Suggesting that \ac{pos} tags and word classes are effective features for reordering. The learnt method is found to be detrimental to the ordering of the \ac{spo} gloss. We believe this result is due to the lack of large-scale training data. As suggested by \citet{lin2020pre} 15 million parallel examples are needed for learnt methods to start outperforming statistical methods. 

Row 1 of both tables shows the BLEU scores between the ground truth gloss and the \ac{spo} gloss, which gives an indication of the performance if the \ac{gs} was 100\% accurate. \Cref{tab:PHIX_reordering} shows a high BLEU-4 score of 37.43, which is the reordering score if we do not apply the \ac{gr} mapping.  Therefore, the \ac{gs} output has the potential to generate a valid translation. Additionally, a proportion of the Deaf community are familiar with \ac{spo} \citep{lucas2014american}, whilst some may even prefer the \ac{spo}. However, further research is required to ascertain whether \ac{spo} translation is useful for the community.

\subsection{State-Of-The-Art Comparison}

The end-to-end \ac{snr} approach joins the output from the \ac{gs} model and the mapping, \(M()\), from \ac{gr} to produce a full translation e.g. \(p(Y|X) = M(p(Y^{SPO}|X))\). We used the mapping from the statistical approach as it was shown to give the best performance in \cref{sec:reodering}. In \cref{tab:phix_comparison} (\ac{ph14t}) and \ref{tab:mdgs_comparison} (\ac{mdgs}) we compare our \ac{snr} approach to state-of-the-art work. Note we can only compare scores that are publicly available, therefore '-' denotes where the authors did not provide results.

For comparison on \ac{mdgs}, we train a \ac{ttg} transformer that achieves a competitive BLEU-4 score compared to \cite{saunders2021signing}. The model is trained till convergence with a beam size of 5 and a word level tokenizer. 

Our results show that reordering is beneficial to the \ac{gs} model, increasing the BLEU-4 score by 1.23 and 1.11 on the \ac{ph14t} Dev and Test sets respectively. On the \ac{mdgs} dataset the reordering mapping was found to only benefit the dev set, increasing the BLEU-4 by 1.4, whilst being detrimental to the test set, decreasing the BLEU-4 by 1.25. The reordering performance is significantly reduced when applied to the output of the \ac{gs} model, decreasing from the theoretical maximum of 53.44 to 19.07 on \ac{ph14t}. We argue this is due to the number of false positives and false negatives in the output of the \ac{gs} model. 

As can be seen from \cref{tab:phix_comparison} and \ref{tab:mdgs_comparison} our models outperformed all other methods on BLEU-1 score \citep{li2021transcribing,saunders2020progressive,saunders2021signing,stoll2018sign}, setting a new state-of-the-art BLEU-1 on \ac{ph14t} and \ac{mdgs}, with a 12.65\% and 37.88\% improvement, respectively. We find our approach outperforms a neural editing program \citep{li2021transcribing}, RNN \citep{stoll2018sign} and a basic transformer \citep{saunders2021signing} on BLEU-1 to 2 and Rouge scores on \ac{ph14t}. While on \ac{mdgs} our approach outperforms a traditional transformer on all metrics. 

\subsection{Model Latency}
\input{Tables/speedup}

\Cref{tab:speedup} demonstrates the significant advantages of our \ac{snr} model. It achieves an impressive 3.08 times speedup when compared to a traditional transformer architecture. Both \ac{gs} and \ac{gr} models utilize the same \ac{nar} decoder, but the incorporation of a reordering mask results in increased latency for the \ac{gr} model. In contrast, our \ac{gs} model, which has shown strong translation performance on its own, exhibits a large speed increase of 18.32 times. These findings highlight the practical utility of our approach, particularly in computationally constrained scenarios.

\input{Tables/state_of_the_art_comparision}

\subsection{Qualitative Experiments}
\label{sec:qualitative}

\begin{figure}[!htb]
    \centering
    \includegraphics[width=0.45\textwidth]{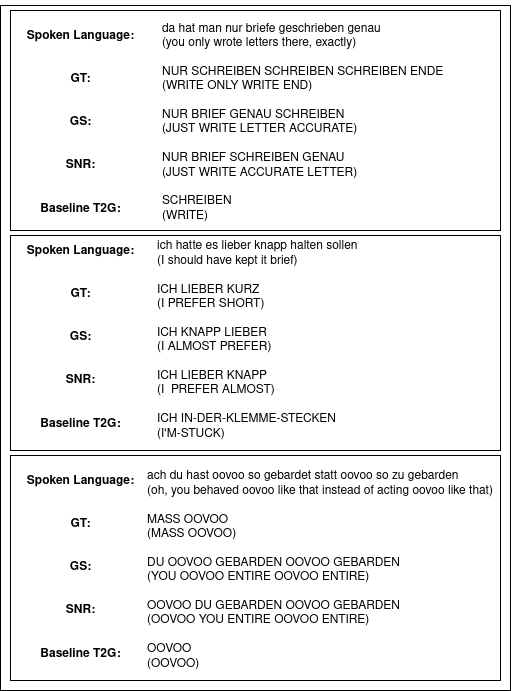}
    \caption{Example \ac{mdgs} translations from a baseline transformer, the \ac{gs} and \ac{snr} models )}
    \label{fig:translation examples}
\end{figure}

\Cref{fig:translation examples} shows example translations from the \ac{mdgs} test set. We compare our approach to the baseline transformer that achieved 31.12 BLEU-1 and 3.04 BLEU-4.  We show the output from the \ac{snr} approach as well as the intermediate output from \ac{gs}.

 These translations show that our approach is better at retaining the meaning of the spoken language sentence, likely due to the 37.88\% improvement in BLUE-1 score. However, in some cases, \ac{gs} can over predict the number of tokens, especially for long input sequences as shown by the third example.

%% file: Tables/speedup.tex
\begin{table}[b]
\centering
\resizebox{0.8\linewidth}{!}{%

\begin{tabular}{r|c|c} 
\toprule
\multicolumn{1}{c|}{Model } & Latency  & Speedup   \\ 
\midrule
T2G Transformer             & 4380ms   & 1.00x     \\
\ac{gs}                     & 239ms    & 18.32x    \\
\ac{gr}                     & 1181ms   & 3.71x     \\
\ac{snr}                    & 1420ms   & 3.08x     \\
\bottomrule
\end{tabular}
}
\caption{Inference latency comparison on \ac{mdgs}.}
\label{tab:speedup}
\end{table}

%% file: Tables/state_of_the_art_comparision.tex
\begin{table*}
\centering
\resizebox{0.9\linewidth}{!}{%
\begin{tabular}{r|ccccc|ccccc} 
\toprule
\multicolumn{1}{l}{\ac{ph14t}}       & \multicolumn{5}{c}{DEV SET}               & \multicolumn{5}{c}{TEST SET}               \\
\multicolumn{1}{c|}{Approach:} & BLEU-1 & BLEU-2 & BLEU-3 & BLEU-4 & ROUGE & BLEU-1 & BLEU-2 & BLEU-3 & BLEU-4 & ROUGE  \\ 
\midrule

T2G \citet{stoll2018sign}                   & 50.15  & 32.47  & 22.30  & 16.34  & 48.42 & 50.67  & 32.25  & 21.54  & 15.26  & 48.10  \\
T2G \citet{saunders2020progressive}         & 55.65  & 38.21  & 27.36  & 20.23  & 55.41 & 55.18  & 37.10  & 26.24  & 19.10  & 54.55  \\
T2G \citet{li2021transcribing}              & -      & -      & 25.51  & 18.89  & 49.91 & -      & -      & -      & -      & -      \\

                                            &        &        &        &        &       &        &        &        &        &        \\
\textbf{\ac{gs}}                            & 62.69  & 38.86  & 25.67  & 17.84  & 56.37 & 60.13  & 35.15  & 21.84  & 14.49  & 54.60  \\

\textbf{\ac{snr}}                           & 62.69  & 40.01  & 27.07  & 19.07  & 56.83 & 60.13  & 35.10  & 22.65  & 15.60  & 53.78  \\
\bottomrule
\end{tabular}
}
\caption{Baseline comparison results for \acf{ttg} translation on the \ac{ph14t} dataset.}
\label{tab:phix_comparison}
\end{table*}

\begin{table*}
\centering
\resizebox{0.9\linewidth}{!}{%
\begin{tabular}{r|ccccc|ccccc} 
\toprule
\multicolumn{1}{l}{\ac{mdgs}}       & \multicolumn{5}{c}{DEV SET}               & \multicolumn{5}{c}{TEST SET}               \\
\multicolumn{1}{c|}{Approach:} & BLEU-1 & BLEU-2 & BLEU-3 & BLEU-4 & ROUGE & BLEU-1 & BLEU-2 & BLEU-3 & BLEU-4 & ROUGE  \\ 
\midrule
T2G  \citet{saunders2021signing}            & -      & -      & -      & 3.17   & 32.93 & -      & -      & -      & 3.08   & 32.52  \\

T2G transformer                             & 31.12  & 14.32  & 6.49   & 3.04   & 34.71 & 31.25  & 15.08   & 7.26  & 3.38  & 34.98  \\
                                            &        &        &        &        &       &        &        &        &        &        \\

\textbf{\ac{gs}}                            & 42.91  & 20.51  & 9.86   & 4.95   & 40.63 & 43.06  & 20.97  & 10.48  & 5.39   & 40.60  \\

\textbf{\ac{snr}}                           & 42.91  & 22.37  & 11.77  & 6.35   & 41.74 & 43.06  & 19.46  & 8.88   & 4.14   & 39.40  \\
\bottomrule
\end{tabular}
}
\caption{Baseline comparison results for \acf{ttg} translation on the \acf{mdgs} dataset.}
\label{tab:mdgs_comparison}
\end{table*}

%% file: Sections/6_Conclusion.tex
In this paper we presented \acf{snr}, a novel two step approach to \ac{ttg} translation, splitting the problem into two concurrent tasks of \acf{gs} and \acf{gr}. This approach disentangles the order from the vocabulary, allowing the \ac{gs} model to focus on maximizing the correct vocabulary whilst leaving arguably the more difficult ordering task to a separate model. We showed our proposed \ac{gs} model achieves a significant increase in BLEU-1 score of 11.79 on the \ac{mdgs} dataset. In addition, we showed that reordering can be learnt by a \ac{gr} model, but statistical based methods perform stronger with the current data limitations. Finally, we showed the result of combining the \ac{gs} with the statistical reordering mapping, finding the \ac{snr} approach outperformed a neural editing program \citep{li2021transcribing}, RNN \citep{stoll2018sign} and a basic transformer \citep{saunders2021signing}.

It's clear that one major challenge to the field is the lack of quality gloss labelled data. Therefore, in the future it would be interesting to see if data augmentation could be used to pool sign language resources from different languages (e.g. DGS, BSL and ASL). A shared lexicon would need to be established across the datasets to combine all of the parallel bilingual data. Alternatively, using the proposed alignment a multilingual model could be trained using Randomly Aligned Substitutions \citep{lin2020pre}.